\pdfoutput=1

\documentclass[11pt]{article}

\usepackage[final]{acl}

\usepackage{times}
\usepackage{latexsym}
\usepackage{amsmath}
\usepackage{amsthm}
\usepackage{enumerate}
\usepackage{fancyhdr}
\usepackage{algorithm}
\usepackage{algpseudocode}
\usepackage{bbm}
\usepackage{marvosym}
\usepackage{array, multirow, boldline}
\usepackage{booktabs}
\usepackage{graphicx}
\usepackage{booktabs}
\usepackage{listings}
\usepackage{amsfonts} 
\usepackage{hyperref}
\usepackage{enumitem}
\usepackage{soul}
\usepackage{authblk}

\usepackage[T1]{fontenc}

\usepackage[utf8]{inputenc}

\usepackage{microtype}

\usepackage{inconsolata}

\usepackage{caption}
\title{SegMix: A Simple Structure-Aware Data Augmentation Method}

\author[1]{Yuxin Pei}
\author[ ]{Pushkar Bhuse\textsuperscript{2}\thanks{pbhuse@ucsd.edu}}
\author[13]{Zhengzhong Liu}
\author[134]{Eric Xing}

\affil[1]{Carnegie Mellon University}
\affil[2]{University of California San Diego}
\affil[3]{Petuum Inc.}
\affil[4]{MBZUAI}

\begin{document}
\maketitle

\begin{abstract}

Interpolation-based Data Augmentation (DA) methods (Mixup) linearly interpolate the inputs and labels of two or more training examples. Mixup has more recently been adapted to the field of Natural Language Processing (NLP), mainly for sequence labeling tasks. However, such a simple adoption yields mixed or unstable improvements over the baseline models.  
We argue that the direct-adoption methods do not account for structures in NLP tasks. To this end, we propose \textbf{SegMix}, a collection of interpolation-based DA algorithms that can adapt to task-specific structures. SegMix poses fewer constraints on data structures, is robust to various hyperparameter settings, applies to more task settings, and adds little computational overhead. In the algorithm's core, we apply interpolation methods on task-specific meaningful segments, in contrast to applying them on sequences as in prior work.
We find SegMix to be a flexible framework that combines rule-based DA methods with interpolation-based methods, creating interesting mixtures of DA techniques. We show that SegMix consistently improves performance over strong baseline models in Named Entity Recognition (NER) and Relation Extraction (RE) tasks, especially under data-scarce settings. Furthermore, this method is easy to implement and adds negligible training overhead.

\end{abstract}

\begin{figure}[tp]
\begin{center}
\includegraphics[width=2.4in]{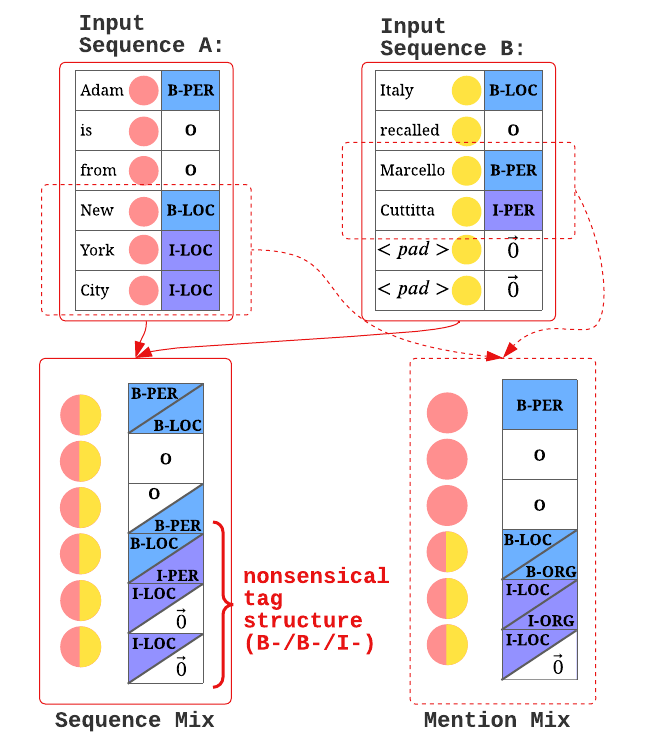}
\caption{
Example of SegMix v.s. Whole-sequence Mixup for NER. 
Each colored block is an entity.
}
\label{fig:seg vs seq}
\end{center}
\end{figure}

\begin{figure*}
\begin{center}
\includegraphics[width=4.4in]{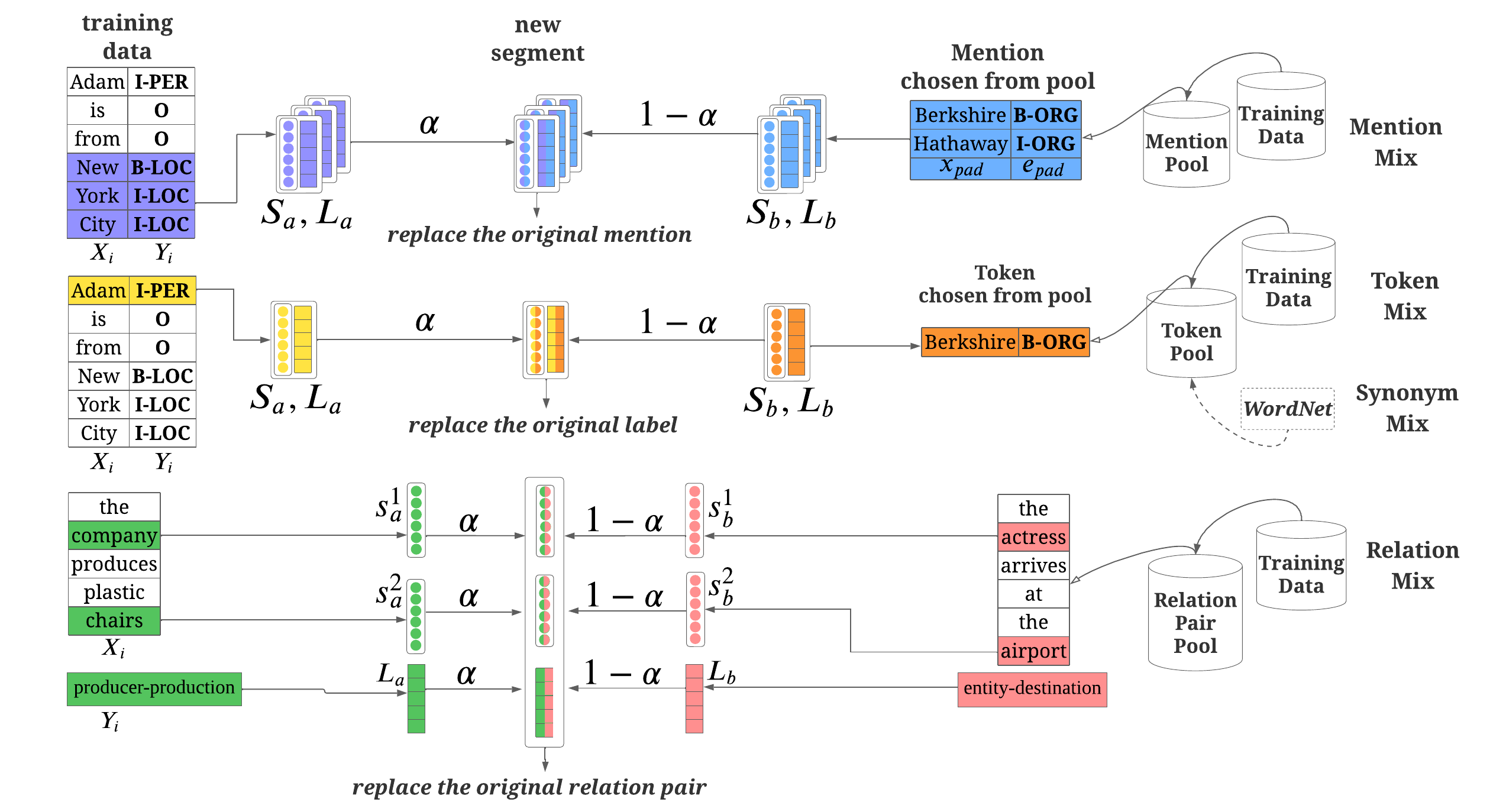}
\caption{Four variations of SegMix (MMix, TMix, SMix, and RMix). The left is the original training sequence. The colored blocks are the segments to be mixed. The segments on the right are randomly sampled from the predefined Segment Pool. Mention Pool, Token Pool, and Relation Pair Pool are constructed from the training data, while the Synonym-token Pool is constructed with the WordNet \citep{10.1145/219717.219748} and returns a synonym of the chosen token. The segment embeddings and one-hot encodings of labels are mixed with ratio $\alpha$.}
\label{fig:MentionMix}
\end{center}
\vspace{-2mm}
\end{figure*} 

\section{Introduction}

Initially proposed as \textit{Mixup} for computer vision tasks, interpolation-based Data Augmentation (DA)~\citep{zhang2018mixup} linearly interpolates the inputs and labels of two or more training examples. 
Inspired by \textit{Mixup}, several attempts have been made to apply interpolation-based DA to NLP, mainly in sequence labeling tasks \citep{guo-etal-2020-sequence}. However, the proposed embedding-mix solution does not extend well to tasks with structured labels. For example, mixing two sentences with different structures usually generates a non-sensical output. As demonstrated in Fig.~\ref{fig:seg vs seq}, when working with entity spans, Whole-sequence Mixup\footnote{\citealt{guo-etal-2020-sequence} is referred to as Whole-sequence Mixup to avoid confusion with SeqMix of \citealt{activeaug}.
} produces non-sensical entity labels like a mixture of nonentity and entity ([O/B-PER]) and consecutive beginning labels ([O/B-PER], [B-LOC/I-PER]). Such noisy augmented data tend to mislead the model, especially in data-scarce settings. As shown in \newcite{lada}, without additional constraints on the augmented data, applying Whole-Sequence Mixup results in performance worse than baseline. 

Instead of using extra heuristic constraints to filter out low-quality augmented data, it may be more efficient and effective to bring structure awareness into the mixing process from the beginning. To this end, we propose \textbf{Segment Mix (SegMix)}, a DA method that performs linear interpolations on meaningful, task-specific segments. Virtuous training examples are created by replacing the original segments with the interpolation of pairs of segment embeddings. As in Fig. \ref{fig:seg vs seq}, the embedding of a location entity (``New York City'') is mixed with the embedding of a person entity (``Marcello Cuttitta''). We exploit the benefit of linear interpolation while keeping the target structure more sensible.

Furthermore, SegMix imposes few restrictions on the original tasks, mixing pairs, or generated examples. On the one hand, this potentially allows one to explore a much larger data space. For example, it allows mixing training samples with various sentence lengths and structures. On the other, it means that SegMix can be applied to other NLP tasks in addition to sequence labeling. 

This paper tests SegMix against Named Entity Recognition (NER) and Relation Extraction (RE), two typical Information Extraction tasks with text segments. We show that SegMix improves upon the baselines under data-scarce settings, and demonstrate its robustness under different hyperparameter settings, which is not the case for simple sequence-based Mixup methods. SegMix is easy to implement\footnote{We will release the experiment code base.} and adds little computational overhead to training and inference. 

\section{Related Work}
Many NLP tasks involve dealing with data with structures, while a popular area is structured prediction. These tasks often involve extracting a predefined target structure from the input data~\citep{10.5555/645530.655813, collins-2002-discriminative, DBLP:journals/corr/MaH16}. NER aims to locate and classify the named entities mentioned in unstructured text. There have been several attempts to apply algorithms similar to \textit{Mixup} to sequence labeling tasks such as NER~\citep{lada, activeaug}. These tasks have linear structures that allow for simple sequence-level mixing methods. RE aims to detect the semantic relationship between a pair of nominals. Unlike NER, RE models typically do not use a linear encoding scheme such as BIO, making sequence-level mixing non-trivial. To the best of our knowledge, interpolation-based DA methods have not been applied to such tasks.

\paragraph{Rule-based DA}
Rule-based DA specifies rules for inserting, deleting, or replacing parts of text~\citep{dataaugmentation}. Easy Data Augmentation {(EDA)}~\citep{wei-zou-2019-eda} proposed a set of token-level random perturbation operations~(insertion, deletion, and swap). SwitchOut~\citep{wang-etal-2018-switchout} randomly replaces tokens in the sentence with random words. WordDrop~\citep{sennrich-etal-2016-edinburgh} drops tokens randomly. Existing work also brings structure awareness into DA. Substructure Substitution (SUB)~\citep{SubstructureSubstitution} generates new examples by replacing substructures (e.g., subtrees or subsequences) with ones with the same label. SUB applies to POS tagging, parsing, and token classification.  A similar idea is proposed for NER~\citep{neraug}. Mention Replacement~(MR) and Label-wise Token Replacement~(LwTR) substitute entity mention and token with those with the same label. Synonym Replacement~(SR) replaces token with a synonym retrieved from WordNet~\citep{WordNet}. \citealt{xu-etal-2016-improved} reverses dependency sub-paths and their corresponding relationships in relation classification. \citealt{sahin-steedman-2018-data} crops and rotates the dependency trees for POS tagging. ~\citealt{https://doi.org/10.48550/arxiv.2104.13913} presents a contrastive pre-training method to  create more generalized representations for RE tasks. It introduces a DA technique where text contained in the shortest dependency path is kept constant and other tokens are replaced. 
Generally, these methods explore the vicinity area around the data point and assume that they share the same label. 

\paragraph{Interpolation-based DA}
Originally proposed for image classification tasks, \textit{Mixup} \cite{zhang2018mixup} performs convex combinations between a pair of data points and their labels. \textit{Mixup} improves the performance of image classification tasks by regularizing the neural network to favor simple linear behavior between training examples \cite{zhang2018mixup}. Several adaptations of \textit{Mixup} have been made in NLP tasks. TMix~\citep{chen2020mixtext} performs an interpolation of text in a hidden space in text classification tasks. Snippext~\citep{Snippext} mixes BERT encodings and passes them through a classification layer for sentiment analysis tasks. AdvAug~\citep{advaug} mixes adversarial examples as an adversarial augmentation method for Neural Machine Translation. 

However, direct application of Whole-Sequence Mixup yields limited improvement in tasks involving structured data. As empirically shown in LADA~\citep{lada} on NER, the direct mixing of two sentences changes both the local token representation and the context embeddings required to identify the entity mention~\citep{lada}. This is also demonstrated in Fig.~\ref{fig:seg vs seq}, the generated data can sometimes be too noisy to help with model training. In fact, LADA has to add additional constraints by mixing the sequences only with its k-nearest neighbors to reduce the noise~\citep{lada}. Similarly, SeqMix~\citep{activeaug} scans both sequences with a fixed-length sliding window and mixes the subsequence within the windows. However, this approach does not eliminate the problem of generating low-quality data --- extra constraints are still used to ensure the quality of generated data. These constraints limit the explorable data space close to the training data. What is more, they complicate the algorithms and add non-negligible computational overheads.

\section{Method}

We propose SegMix and implements 4 variants, namely MentionMix (MMix), TokenMix (TMix), SynonymMix (SMix), and RelationMix (RMix). As shown in Fig. \ref{fig:MentionMix}, after defining the task-dependent segment, we create a new training sample by replacing a segment of the original sample with a mixed embedding of the segment itself and another randomly drawn segment. These mixed embeddings are then fed into the encoder. Algorithm~\ref{algo:segmix} presents the SegMix generation process.

\begin{algorithm}
\caption{SegMix generation algorithm}
\label{algo:segmix}
\begin{algorithmic}[1]
\State \textbf{Input:} $\mathcal{D}, \mathcal{P}^k, r$
\State $\mathcal{D}_A\gets \{\}, \mathcal{D}_S \gets $ sample($\mathcal{D}, \emph{len}(\mathcal{D})\cdot r$)\label{algo:line:2}
\For{$(X_i, Y_i)$ in $\mathcal{D}_S$}\label{algo:line:3}
\State $E_i, O_i \gets \textbf{Emb}(X_i), \textbf{OHE}(Y_i)$\label{algo:line:4}
\State $\lambda \gets Beta(\alpha, \alpha)$\label{algo:line:5}
\State $S_a, l_a \gets$ $k$ segment tuples in $X_i, Y_i$ \label{algo:line:6}
\State $S_b, l_b \gets$ $k$ segment tuples in $\mathcal{P}$\label{algo:line:7}
\State $X_i', Y_i' \gets X_i.$\lstinline{copy()}, $Y_i.$\lstinline{copy()}
\For{$s_a^j, s_b^j$ in $S_a, S_b$}
\State $e_a, e_b = \textbf{Emb}(s_a), \textbf{Emb}(s_b)$
\State $start, end \gets $ index range of $s_a^j$ in $X_i$
\State $\tilde{e}_a^j, \tilde{e}_b^j \gets $pad\_to\_longer($e_a^j, e_b^j$)\label{algo:line:12}
\State $E_i[start: end] \gets \tilde{e}_a^j\cdot \lambda + \tilde{e}_b^j\cdot (1-\lambda)$\label{algo:line:13}
\EndFor
\For{$l_a^j, l_b^j$ in $l_a, l_b$}
\State $o_a, o_b = \textbf{OHE}(l_a), \textbf{OHE}(l_b)$
\State $start, end \gets $ index range of $l_a^j$ in $Y_i$
\State $\tilde{o}_a^j, \tilde{o}_b^j \gets $pad\_to\_longer($o_a^j, o_b^j$)\label{algo:line:18}
\State $O_i[start: end] \gets \tilde{o}_a^j\cdot \lambda + \tilde{o}_b^j\cdot (1-\lambda)$\label{algo:line:19}
\EndFor
\State $\mathcal{D}_A$.add($(E_i, O_i)$)
\EndFor
\State \textbf{Output}: $\mathcal{D}_A$
\end{algorithmic}
\end{algorithm}

Formally, consider a training dataset $\mathcal{D} = \{(X_i, Y_i) | i \in N\}$  of size $N$, where each input $X_i$ is a sequence of tokens $X_i=(X_i^1, X_i^2, \dotsc, )$ and a task-dependent structured output $Y_i$, a structured prediction algorithm generally encodes the output $Y_i$ using a task-dependent scheme. For example, NER labels are often encoded with the BIO scheme while RE labels are associated with a pair of nominal phrases. SegMix adapts to different encoding schemes by designing task-dependent segments.

A segment $s(u,v)$ is a continuous sequence of tokens $(X_i^u, X_i^{u+1}, \dotsc, X_i^v)$ in sample $X_i$, a segment tuple $S = [s_i(u_i,v_i),...]$ is a $k-$ary tuple of segments contained in the sequence. We choose a segment tuple relevant to the task and associate it with an appropriate label list $L = [l_i,...]$. For example, in RE, there are segment tuple of length 2, which contains the pair of nominals in a relation.

A Segment Pool of size $M$:$\mathcal{P}^k = \{(S_i, L_i) | i \in M\}$ is generated by collecting segment tuples $S_i$ from the training data or an external resource (e.g. \textit{WordNet}). Here, $k$ is a constant for a specific task. For example, in RE, there are binary segment tuple containing a pair of nominals.

With the training data set $\mathcal{D}$, the Segment Pool $\mathcal{P}^k$, and the mix rate $r$, \texttt{SegMix}~$(\mathcal{D}, \mathcal{P}^k, r)$ returns an augmented data set $\mathcal{D}_A$ of size $r\cdot N$. 
A set $\mathcal{D_S}$ of size $r\cdot N$ is first drawn from the training data $\mathcal{D}$ as candidates for augmentation.
For each data point $(X_i, Y_i)$ drawn from $\mathcal{D_S}$, we randomly pick a segment tuple $S_a$ and the corresponding label list $L_a$ from the sequence $X_i$. The mix for candidate $X_i$, $(S_b, L_b)$, is then drawn from the Segment Pool.

Let $\mathbf{Emb}$ be an embedding function on $\mathbb{R}^V \mapsto \mathbb{R}^D$, where V is the size of the vocabulary and D is the embedding dimension. Let $\mathbf{OHE}$ be a function that returns the one-hot encoding of a label.

For all $s_a, s_b = S_a[i], S_b[i], 1 \leq i \leq \lstinline{len}(S_a)$, and $l_a, l_b =L_a[j], L_b[j], 1 \leq j \leq \lstinline{len}(L_a)$. Define $e_a, e_b=\mathbf{Emb}(s_a), \mathbf{Emb}(s_b), o_a, o_b=\mathbf{OHE}(l_a), \mathbf{OHE}(l_b)$.

The embeddings and one-hot encodings are then padded according to sequence length (line \ref{algo:line:12}, \ref{algo:line:18}). Let $\tilde{e}_a, \tilde{e}_b, \tilde{o}_a, \tilde{o}_b$ be the padded version of the embeddings and one-hot encodings. Finally, in line \ref{algo:line:13}, \ref{algo:line:19}, we perform a linear interpolation between $\tilde{e}_a, \tilde{e}_b$ and $\tilde{o}_a, \tilde{o}_a$  with a mix rate $\lambda$ chosen randomly from a Beta distribution~(see specifications in \ref{subsec:impledetails}):
\begin{equation} 
\begin{aligned}
&e_a' \gets \tilde{e}_a \cdot \lambda + \tilde{e}_b\cdot (1-\lambda) \\
&o_a' \gets \tilde{o}_a \cdot \lambda + \tilde{o}_b\cdot (1-\lambda) 
\end{aligned}
\label{eq1} 
\end{equation}
In Eq.\ref{eq1}, $\cdot$ is a scalar multiplication and $+, -$ are vector element-wise operations. When $\lambda=1$, the augmented data falls back to the original one. When $\lambda=0$, the segments are completely replaced by those drawn from the pool, equivalent to replacement-based DA techniques.

Finally, the augmented data point is generated by copying the original data and replacing the chosen segment and labels with the mixed version. We present 3 variations of SegMix for NER and 1 for RE with different types of Segment Pool $\mathcal{P}^k$.

\paragraph{MentionMix} Inspired by MR, MMix performs linear interpolations on a mention level (a contiguous segment of tokens with the same entity label). A Mention Pool $\mathcal{P}^1$ is constructed by scanning the training data set and extracting all mention segments and their corresponding labels. Thus, each segment tuple is composed of a single mention and a list of entity labels encoded with the BIO scheme. This method can also be viewed as a generalization of (SUB)~\citep{SubstructureSubstitution}
which  performs a soft-mix of substructures of varying lengths.


\paragraph{TokenMix} Inspired by LwTR, TMix performs linear interpolations at the token level. We use tokens with entity labels in the BIO scheme of training data sets as a token pool $\mathcal{P}^1$. Each segment tuple is composed of a single token and its label.

\paragraph{SynonymMix} Inspired by SR, the Synonym Pool $\mathcal{P}^1$ returns a synonym of the token in the original sequence based on \textit{WordNet}~\citep{WordNet}. We assume the two synonyms share the same label, thus interpolation only happens within input.

\paragraph{RelationMix} Since each relation is composed of two possibly nonadjacent nominals in a sentence, we construct a pool $\mathcal{P}^2$ with groups of two nominals and a relation label\footnote{The direction of the relation is implied by the labels. For example, the label list contains both producer-product (e1,e2) and producer-product (e2,e1)}. During the mixing phase, the two nominals and their corresponding relation labels are mixed with a pair of nominals from $\mathcal{P}^2$.

\section{Experiments}


\paragraph{Datasets}

\begin{table}[h]
\centering
\begin{tabular}{l ccc}\toprule
  & Language & Task & \# Instances  \\ \midrule
 CoNLL-03& English & NER & \small 14987 \normalsize \\
 \textit{Kin} & Kinyarwanda & NER & \small 626 \normalsize \\
 \textit{Sin} & Sinhala & NER & \small 753 \normalsize \\
 SemEval & English & RE & \small 8000 \normalsize\\
 DDI & English & RE & \small 22233 \normalsize\\
 Chemport & English & RE & \small 18035 \normalsize\\
\bottomrule
\end{tabular}
\caption{Dataset Statistics}
\label{table:dataset}
\end{table}

We conduct SegMix experiments mainly on $3$ datasets for NER and $3$ for RE on a variety of domains and languages. An NER task is to recognize mentions from text belonging to predefined semantic types, such as person, location, and organization. An RE task requires one to classify the relation type between two prelabeled nominals in a sentence. Some basic dataset statistics are included in Table.~\ref{table:dataset}\footnote{Since no down-sampling settings are included in LORELEI-Kin and Sin, we report the results as a single value.}.


\begin{enumerate}[label={(\arabic*)},noitemsep,nolistsep]
    \item CoNLL-03~\citep{DBLP:journals/corr/cs-CL-0306050}, an English corpus for NER containing entity labels such as person, location, organization, etc.\footnote{We also conduct experiments on GermEval, a German NER dataset. The results and trends are similar to those in CoNLL-03, and are presented in the Appendix. A.1} 
    \item LORELEI~\citep{strassel-tracey-2016-lorelei} which contains NER annotations for text in languages Kinyarwanda (\textit{Kin}) and Sinhala (\textit{Sin}).
    \item SemEval-2010 Task 8~\citep{semeval}, an English corpus for RE task, containing $9$ relation types that include cause-effect, product-producer, instrument-agency, etc.
    \item DDI~\citep{DDI}, a biomedical dataset manually annotated with drug-drug interactions, containing $4$ relationship types.
    \item ChemProt~\citep{Krallinger2017OverviewOT}, a biomedical dataset annotated with chemical-protein interactions, containing $4$ interaction types.
\end{enumerate}

\begin{table*}
\centering
\small
\begin{tabular}{l ccc c c}\toprule
& \multicolumn{3}{c}{CoNLL-03} & Kin & Sin
\\\cmidrule(lr){2-4}
Data Size  & 200 & 400 & 800 & 626 & 753 \\\midrule
BERT & $76.03$ \text{\small $\pm$} $0.57$ & $81.20$ \text{\small $\pm$}  $0.29$ & $84.34$ \text{\small $\pm$}  $0.33$ & $82.29$ & $75.02$  \\ \midrule
BERT + LADA & $70.46$ \text{\small $\pm$}  $0.84$ & $81.98$ \text{\small $\pm$}  $0.16$ & $84.53$ \text{\small $\pm$}  $0.09$ & $76.02$ & $60.43$ \\
BERT + SeqMix & $77.10$ \text{\small $\pm$}  $1.04$ & $81.55$ \text{\small $\pm$}  $0.66$ & $84.89$ \text{\small $\pm$}  $0.27$ & $83.13$ & $78.93$\\ 
BERT + Whole-seq Mix & $75.11$ \text{\small $\pm$}  $0.62$ & $81.94$ \text{\small $\pm$}  $0.14$ & $84.61$ \text{\small $\pm$}  $0.18$ & $82.35$ & $79.17$\\ 
BERT + MR  & $77.86$ \text{\small $\pm$}  $0.36$ & $81.49$ \text{\small $\pm$}  $0.17$ & $84.21$ \text{\small $\pm$}  $0.29$ & $83.46$ & $78.62$\\
BERT + LwTR  & $76.69$ \text{\small $\pm$}  $0.49$ & $81.13$ \text{\small $\pm$}  $0.36$ & $84.56$ \text{\small $\pm$}  $0.37$ & $82.42$ & $78.17$\\
BERT + SR  & $77.35$ \text{\small $\pm$}  $0.29$ & $81.33$ \text{\small $\pm$}  $0.32$ & $85.10$ \text{\small $\pm$}  $0.11$ & $82.51$ & $78.38$\\ \midrule
BERT + \textbf{MMix} \textdagger & $78.51$ \text{\small $\pm$}  $0.34$ & $\mathbf{82.98}$ \text{\small $\pm$}  $0.61$ & $85.37$ \text{\small $\pm$}  $0.59$ & $83.37$ & $79.50$ \\ 
BERT + \textbf{TMix} \textdagger & $\mathbf{78.75}$ \text{\small $\pm$}  $0.49$ & $82.28$ \text{\small $\pm$}  $0.30$ & $85.51$ \text{\small $\pm$}  $0.21$ & $\mathbf{83.85}$ & $78.63$ \\ 
BERT + \textbf{SMix} \textdagger & $77.95$ \text{\small $\pm$}  $0.38$ & $82.51$ \text{\small $\pm$}  $0.36$ & $85.33$ \text{\small $\pm$}  $0.19$ & $83.31$ & $79.38$ \\ 
BERT + \textbf{MMix} + \textbf{SMix} \textdagger & $78.45$ \text{\small $\pm$}  $0.26$ & $82.39$ \text{\small $\pm$}  $0.21$ & $85.66$ \text{\small $\pm$}  $0.25$ & $82.81$ & $79.83$  \\ 
BERT + \textbf{MMix} + \textbf{TMix} \textdagger & $78.46$ \text{\small $\pm$}  $0.26$ & $82.39$ \text{\small $\pm$}  $0.24$ & $\mathbf{85.82}$ \text{\small $\pm$}  $0.21$ & $82.75$ & $\mathbf{80.31}$ \\
BERT + \textbf{MMix} + \textbf{SMix} + \textbf{TMix} \textdagger & $78.21$ \text{\small $\pm$}  $0.28$ & $82.36$ \text{\small $\pm$}  $0.34$ & $85.26$ \text{\small $\pm$}  $0.27$ & $82.83$ & $78.05$ \\
\midrule
RoBERTa \textdagger & $74.08$ \text{\small $\pm$}  $0.27$ & $78.89$ \text{\small $\pm$}  $0.59$ & $82.28$ \text{\small $\pm$}  $0.23$ & $-$ & $-$ \\ 
\midrule
RoBERTa +\textbf{MMix} \textdagger & $\mathbf{75.31}$ \text{\small $\pm$}  $0.52$ & $\mathbf{80.09}$ \text{\small $\pm$}  $0.49$ & $83.37$ \text{\small $\pm$}  $0.54$ & $-$ & $-$ \\ 
RoBERTa + \textbf{TMix} \textdagger & $74.55$ \text{\small $\pm$}  $0.37$ & $79.44$ \text{\small $\pm$}  $0.35$ & $83.22$ \text{\small $\pm$}  $0.80$ & $-$ & $-$ \\ 
RoBERTa + \textbf{SMix} \textdagger & $75.18$ \text{\small $\pm$}  $0.42$ & $79.80$ \text{\small $\pm$}  $0.45$ & $\mathbf{83.49}$ \text{\small $\pm$}  $0.39$ & $-$ & $-$ \\ 
\bottomrule

\end{tabular}
\caption{F1 scores for NER in data-scarce settings (downsampled CoNLL-03 and LORELEI (\textit{Kin} and \textit{Sin})) using SegMix compared with interpolation- and replacement-based DA methods. We use $5$ different random seeds for down-sampled datasets and report their averaged performance and standard deviation as $\mu \pm \sigma$\protect. For LORELEI, we report the $10$-fold cross-validation result. Although there is no one best performing variant of SegMix for all settings, we observe that for all variants, SegMix had the best performance compared to the baseline in all settings and other DA techniques in most settings. \textdagger denotes our methods.
}
\label{table:main}
\end{table*}

\paragraph{Data Sampling}
For true low-resource languages Kinyarwanda and Sinhala (data sizes of LORELEI-Sin and LORELEI-Kin are less than $5\%$ of the CoNLL-03 English dataset), we use all available data.
To create difference scarce settings for CoNLL-03, we subsample a range of sizes ($200, 400, 800, 1600, 3200, 6400, 12800$) of the original training data as the training set. The augmentation algorithm can only access the downsampled training set. We use $5$ different random seeds to subsample the training set of each size and report both mean and standard deviation as $(\mu \pm \sigma)$. The validation and test dataset are unchanged. For LORELEI, we deleted all data samples that only have character "--". Therefore, there are some discrepancies between our reported data number and the original paper. For RE, we subsample ($100, 200, 400, 800, 1600, 6400$) from the original training data as the training set. We do not continue experiments for larger sizes since the improvement from DA diminished.

\paragraph{Settings}
For each data split, we conduct experiments on $12$ settings for NER ---- $2$ interpolation-based DA (Inter+Intra LADA\footnote{We used implementation available at \url{https://github.com/GT-SALT/LADA}.}, Whole-sequence Mixup\footnote{Implemented by setting segments as whole sequences.}), $3$ replacement based DA (MR, SR, LwTR)\footnote{Implemented as SegMix where mix rate is 1.}, and $6$ variations of SegMix (MMix, TMix, SMix, and their combinations MMix + SMix, MMix + TMix, MMix + TMix + SMix) with a fixed $0.2$ augmentation rate. We use the BIO tagging scheme~\cite{marquez-etal-2005-semantic} to assign labels to each token in NER tasks. In RE tasks, we compare RMix with Relation Replacement. Gold standard nominal pairs are used.

All the methods are evaluated with F1 scores. For \textit{Kin} and \textit{Sin}, we report the average F1 scores over 10 folds with cross-validation, which is consistent with \citealt{rijhwani-etal-2020-soft}.



\subsection{Implementation Details} \label{subsec:impledetails}
For our experiments, we adopt the pretrained BERT and RoBERTa models\footnote{The model choices are included in Appendix~\ref{subsec:baseline}.} 
as the encoder, and a linear layer to make prediction, with soft cross-entropy loss. The pretrained BERT model is adopted for each language whereas due to computation expenses, we adopted the pretrained RoBERTa model for experiments on only the CoNLL-03 dataset. For pseudo-data-scarce settings (CoNLL-03, DDI, Chemprot, and SemEval), we train all the models for 100 epochs with early stopping and take the checkpoint with the maximum validation score on the development dataset as the final model. For \textit{Kin} and \textit{Sin}, under each data split, we train the model for 100 epochs and report the F1 score. The initial weight decay is 0.1 and $\alpha$ is 8 for both models. Additionally, learning rates for all settings are set to $5e-5$ for the BERT model and $1e-4$ for the RoBERTa model.

\subsection{Results and Analysis}
\begin{figure*}
    \centering
    \includegraphics[width=2.6in]{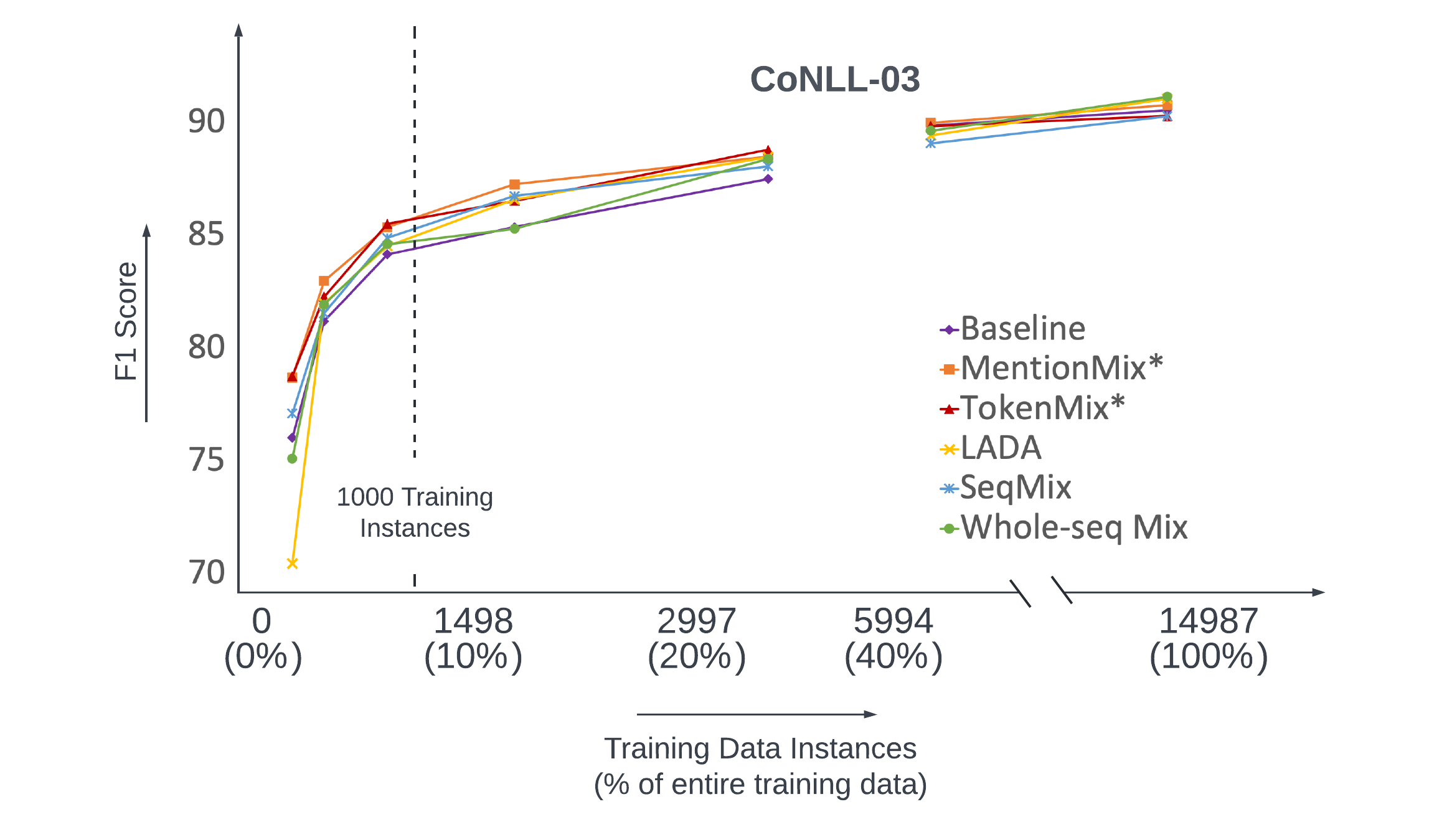}
    \includegraphics[width=2.6in]{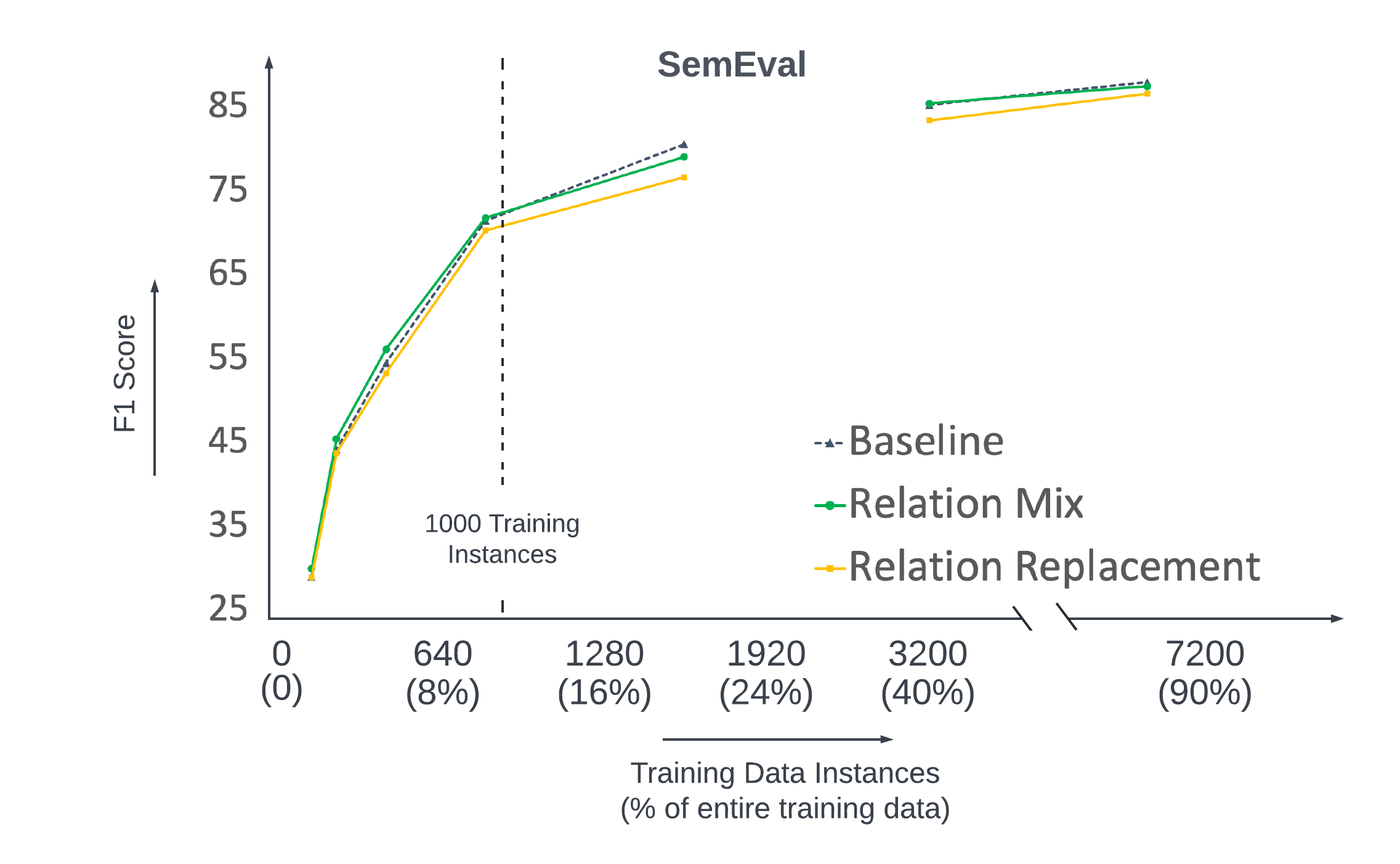}\\
    \includegraphics[width=2.6in]{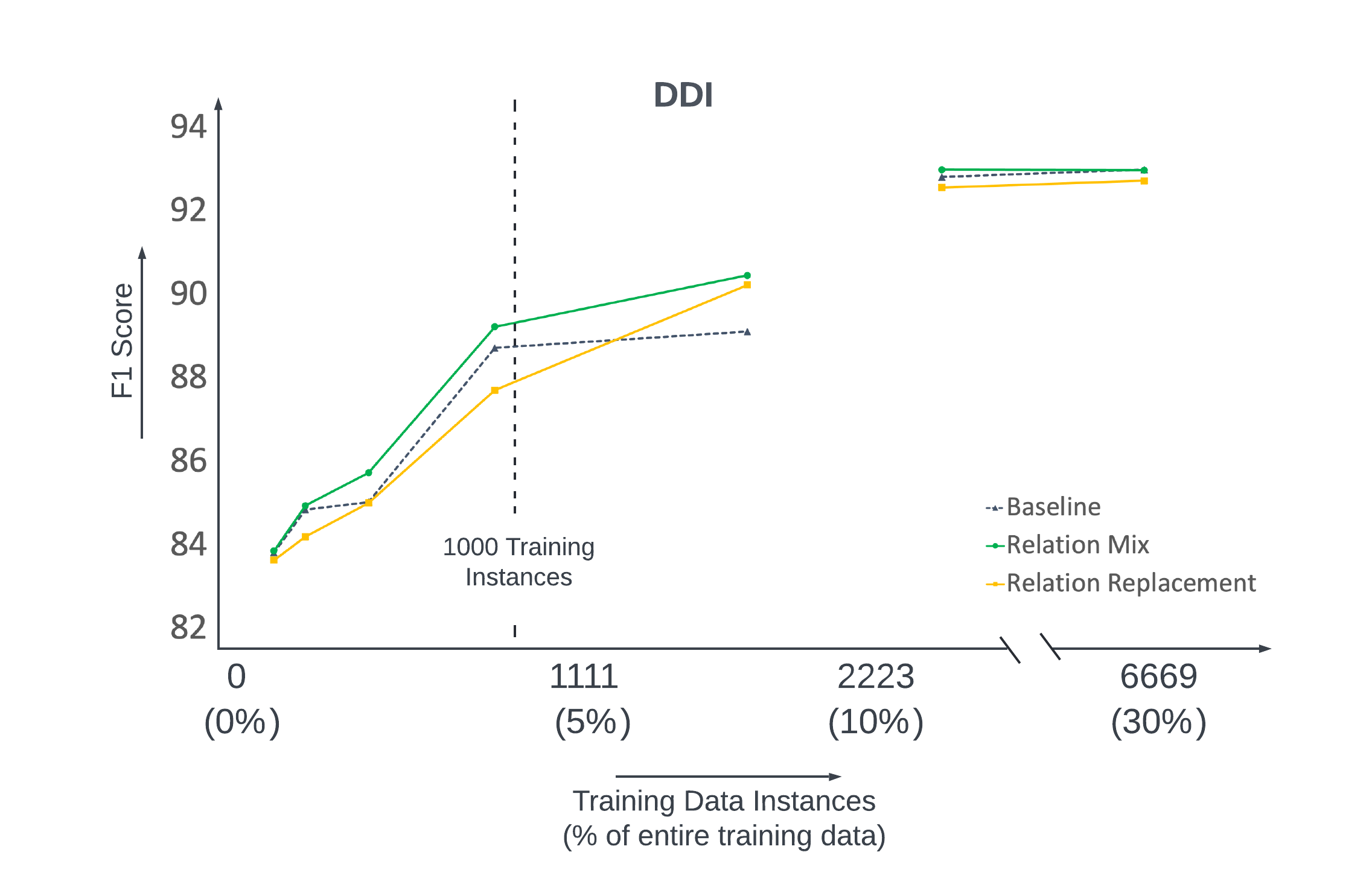}
    \includegraphics[width=2.6in]{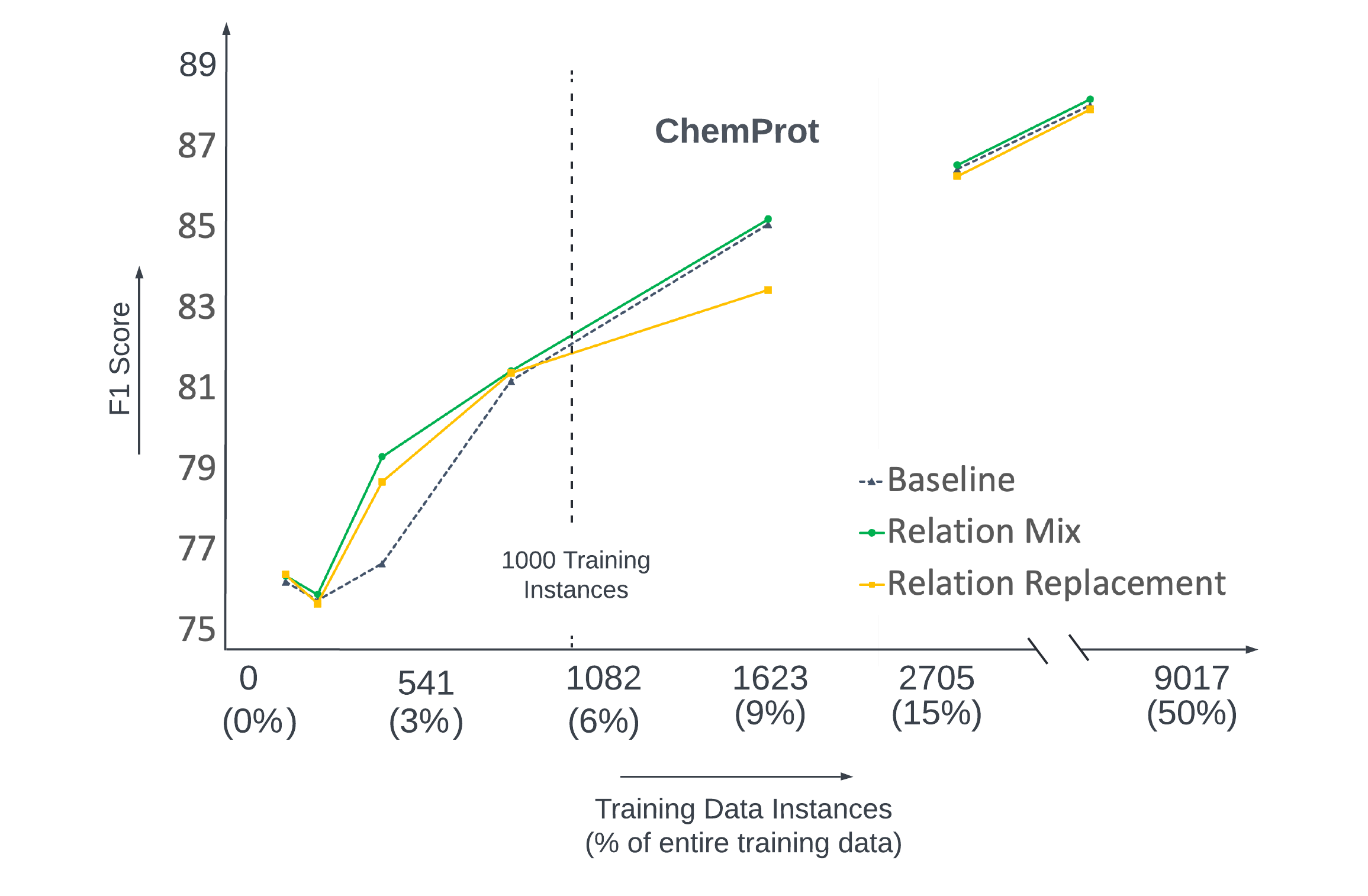}
    \caption{Average F1 score on CoNLL-03, DDI, ChemProt, and SemEval-2010 under different down-sampled data settings. The y axis represents the average F1 score, and the x axis represents number and percentage of instances used as the training set. For each dataset, we calculate the average F1 score on increasing data sub-samples until the performance of our SegMix variant either plateaus or  equals that of the baseline. SegMix works best in settings with less than approximately 1000 training instances.}
    \label{fig:re_res}
\end{figure*}

\paragraph{NER} The results for the three NER datasets under data-scarce settings with BERT and RoBERTa are shown in Table~\ref{table:main}. Fig.~\ref{fig:re_res} includes the results for CoNLL-03 under all data settings with BERT. Under all settings, SegMix or a combination of SegMix achieves the best result compared with other interpolation- and replacement-based methods. For BERT, the best performing SegMix improves the baseline by $2.7$ F1 in CoNLL-03 with the $200$ sample setting, $1.5$ F1 for \textit{Kin}, and $5$ F1 for Sin. As for RoBERTa, SegMix and its variants perform better compared to the baseline RoBERTa model in all simulated data-scarce scenario with CoNLL-03. For example, the best performing SegMix variant with RoBERTa improves the baseline by 1.2 F1 on CoNLL-03 under the $200$-sample setting. SegMix proves to be effective under both down-sampled settings and true low-resource settings. These results are consistent with our hypothesis that the ``soft'' mix of data points in structure-aware segments yields better results than ``hard'' replacement or mixing on sequences.
In comparison, LADA has an unstable performance under data-scarce settings. It produces worse results than the baseline under the CoNLL-03 with $200$ samples, and in both low-resource languages \textit{Kin} and \textit{Sin}, while SegMix shows consistent improvements. 

One notable trend is that most DA methods provides a larger improvement on \textit{Sin} in compared to \textit{Kin}. Notice that even with the same model architecture, the baseline performance of \textit{Sin} is considerably lower compared to the performance of \textit{Kin} and English of similar data sizes. 
This could be due to the fact that multilingual BERT transfers better between languages that share more\footnote{While both the Kinyarwanda-BERT and Sinhala-BERT are transferred from M-BERT, the number of common grammatical ordering WALS features~\cite{wals} is 3 between Kinyarwanda and English and 1 for Sinhala. These features are 81A, 85A, 86A, 87A, 88A and 89A.} word order features~\cite{pires-etal-2019-multilingual}.
Given the lower baseline, many DA methods provide larger improvements in \textit{Sin} compared to \textit{Kin}, and our SegMix variants score around 80 F1 scores. This shows that DA methods are generally very valuable for low resource and understudied languages.


\paragraph{RE}
For RE, we compare RMix with the baseline and Relation Replacement (replacing nominal pairs). The results are presented in Fig.\ref{fig:re_res}. We find that simple replacement sometimes worsens the baseline performance, while RMix consistently improves the baseline. We analyze its  performance on increasing percentages of training data to simulate pseudo-data-scarce settings, as well as settings with ample training data. We observe a consistent improvement performance of RMix over replacement based methods, and at least comparable performance with the baselines. SegMix performs well in data scarce settings, more specifically, on scenarios with less than approximately 1000 training examples. For example, in case of the DDI dataset, SegMix performs at least 2 F1 scores better compared to the baseline in these scenarios.





\begin{figure*}
    \centering
    \includegraphics[width=2.6in]{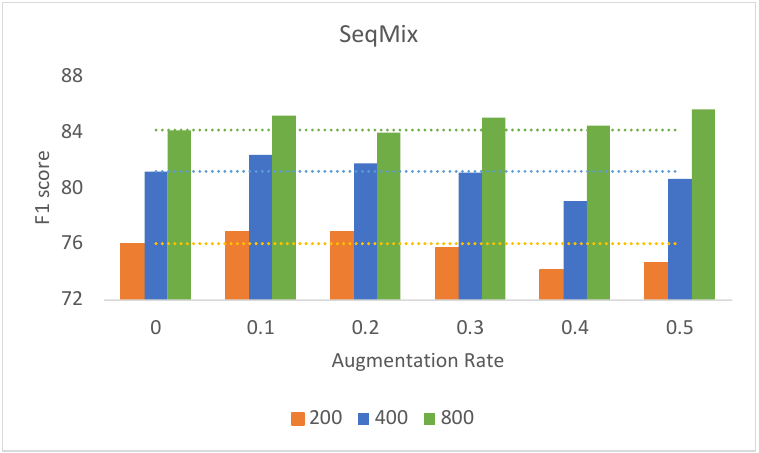}
    \includegraphics[width=2.6in]{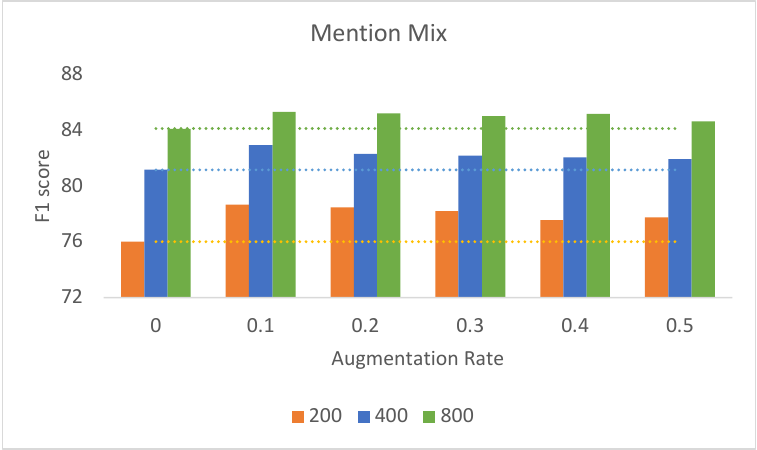}
    \caption{Average F1 score with variant augmentation rates of MMix and SeqMix on CoNLL-03 with 200, 400, and 800 down-sampled data. The colored line represents the baseline performance. MMix constantly outperforms the baseline performance.}
    \label{fig:aug_rate}
\end{figure*}

\paragraph{Robustness with respect to augmentation rate} 
From previous results on sequence-level Mixup~\citep{activeaug, lada}, we observe that the performance of the model tends to drop below the baseline as the augmentation rate increases above a certain value. Furthermore, the optimal augmentation rate varies under different initial data settings: a good augmentation rate for the $200$-sample might not be good for the 800-sample. With BERT, for example, a $0.2$ augmentation rate improves upon baseline under the $200$-sample setting, but produces worse results than the baseline under the $800$-sample setting. This leads to an extra burden in hyperparameter tuning. Through experiments on varying augmentation rates under $3$ different data-scarcity settings, we show that MMix consistently improves the baseline performance under all settings, making it more applicable in practical contexts. As presented in Fig.~\ref{fig:aug_rate}, MMix consistently improves upon the baseline for all experimented augmentation rate. Furthermore, the best performance is consistently achieved at $0.1$. TMix and SMix also show a similar trend, the specific scores are presented in Appendix.~\ref{sec:add_res}.

\paragraph{Computation Time} 
SegMix is easy to implement and adds little computational overhead. We compare the time required to generate the mixing data and training using LADA, MMix, and SeqMix in Table.~\ref{table:comptime}. Without extra constraints on the augmentation process, MMix (and its other variants) takes <$1$ second on average to generate the augmented dataset. While SeqMix takes >$2$ minutes due to the filtering process. Both SeqMix and SegMix pass mixed embeddings into the encoder directly; thus, no extra computation is required for each epoch. However, we observe that SegMix converges faster than SeqMix, thus requiring less training time on average. Since LADA mixes hidden representations during training, no augmented dataset is explicitly generated. This leads to almost twice the training time of SegMix.

\begin{table}
\centering
\begin{tabular}{lcc}\toprule
   & mixing time (s) & training time (s) \\ \midrule
SeqMix & \small$138.90$  $\pm$  $15.46$ \normalsize & \small $1094.99$ $\pm$ $108.28$ \normalsize \\
MMix \textdagger  & \small $0.81$  $\pm$  $0.22$ \normalsize & \small $609.61$ $\pm$  $66.39$ \normalsize \\
LADA & -- & \small $1120.78$ $\pm$  $103.13$ \\\bottomrule
\end{tabular}
\caption{Comparison of the mixing time (time taken to generate the augmented data) and the training time (time taken to train the model to converge) of LADA, SeqMix and MMix on CoNLL-03 with $200$ downsampled data. We experimented with $5$ different random seeds and reported the average time and standard deviation.}
\label{table:comptime}
\end{table}

\subsection{Discussion}
We argue that SegMix
keeps the syntactic and output structure of training data intact. We choose some sample sequences in CoNLL-03 and visualize them in Fig.~\ref{mix_samples} by mapping the mixed embeddings to the nearest word in the vocabulary.

\begin{figure}[ht]
\fbox{
    \small
    \begin{minipage}{0.97\linewidth}
    \underline{Original}: \textbf{Swedish} \text{\scriptsize\textcolor{red}{[MISC]}} options and derivatives exchange \textbf{OM Gruppen AB} \text{\scriptsize\textcolor{red}{[ORG]}} said on Thursday it would open an electronic bourse for forest industry products in \textbf{London} \text{\scriptsize\textcolor{red}{[LOC]}} in the first half of 1997.\\
    \underline{MMix}: \textbf{Swedish} \text{\scriptsize\textcolor{red}{[MISC]}} options and derivatives exchange \textbf{Javier Gomez de} \text{\scriptsize\textcolor{red}{[PER/ORG]}} said on Thursday it would open an electronic bourse for forest industry products in \textbf{London} \text{\scriptsize\textcolor{red}{[LOC]}} in the first half of 1997.\\
    \underline{Whole-Sequence Mix}: \textbf{Sweden} \text{\scriptsize\textcolor{red}{[MISC/ORG]}} \textbf{option} \text{\scriptsize\textcolor{red}{[O/ORG]}} but [unused33] transfer \textbf{. . [unused10]} \text{\scriptsize\textcolor{red}{[O/ORG]}} saying to Friday them might closed his electronics . with woods companies Products of \textbf{Paris} \text{\scriptsize\textcolor{red}{[O/LOC]}} of a second three in 1995.
    \end{minipage}
}
\caption{Mixed sentence samples recovered by mapping embeddings to the nearest token (l2 distance). [A/B] represents the linear interpolation of the one-hot encodings of the two labels A and B.\label{mix_samples}}
\end{figure}

MMix preserves the syntactic and entity structures while achieving linear interpolation between each mention. Due to the high proportion of non-entity phrases in the dataset, SeqMix tends to mix entity mentions with nonentity segments (label [O]). The resulting sentences often contain nonmeaningful entities (e.g., \textit{option} and \textit{. . [unused10]}), but are perceived as entities (with a non-[O] label). The nonentity phrases in the sentence would also be mixed, producing semantically incorrect context phrases like \textit{second three in 1995}.


Unlike other interpolation-based DA methods, SegMix imposes few constraints on the mixing candidate and mixed examples. All training data pairs can potentially be used as mixing candidates and no filtering process is required after the augmented sample is generated. This not only potentially expands the explorable space of our augmentation algorithm but also saves computational time. 

When analyzing the improvement for each entity class for CoNLL-03, there is an overall improvement in the accuracy for each class, especially for PER and ORG\footnote{Confusion Matrix included in Appendix. \ref{sec:add_res}}. Before SegMix, the model tends to mistakenly predict [LOC] for [ORG] ($27\% \rightarrow 19\%$), and [O] for [PER] ($19\% \rightarrow 8\%$). This may be due to the fact that MMix introduces more variations of meaningful entities into the training process, preventing the model from only predicting labels with the one of majority occurrence. 

We also analyze cases that are improved in different tasks, the specifics can be found in Appendix.\ref{subsec:case_analysis}. In one example, the baseline model correctly detects a entity span "British University", but falsely classifies it as [MISC] whereas SegMix correctly distinguishes it as an [ORG]. In another example, the baseline model fails to detect the entity span ("Minor Counties" instead of "Minor Counties XI") and the correct entity while SegMix gives the same wrong span, but correct entity class.
We hypothesize that SegMix mainly helps the model distinguish between ambiguous types instead of span detection. To validate this claim, we convert all mentions to [B] and [I] during the inference phase and find that there is little difference between the models (both around $98\%$) in terms of span accuracy --- confirming our hypothesis. Similarly for RE, we conduct evaluation in two settings: evaluating only relation type and only relation direction. The accuracy scores for the two metrics both increase around $2\%$. Thus, RMix helps to identify both the correct type and direction of relations. Specific cases and examples can be found in Appendix~\ref{subsec:case_analysis}.


\paragraph{Limitations} In this paper, we analyze the efficacy of SegMix on tasks with clear task related segments (NER and RE). SegMix works best in such settings but we do not validate it on tasks like syntactic parsing. Secondly, we only test the performance of SegMix on a few transformer based models (BERT and RoBERTa), it is not applicable to new paradigms such as question answering and generation based information extraction techniques~\cite{he-etal-2015-question,josifoski-etal-2022-genie}. Lastly, although SegMix works best on small datasets ($\approx$1000 examples), we recognize that it has a diminishing improvement with the increase of data size. Thus, we recommend using SegMix in data-scarce situations.

\section{Conclusion}
This paper proposes SegMix, a simple DA technique that adapts to task-specific data structures, which extends the application range of \textit{Mixup} in NLP tasks.
We demonstrate its robustness by evaluating model performance under both true low-resource and downsampled settings on multiple NER and RE datasets. SegMix consistently improves the model performance and is more consistent than other mixing methods. By combining rule-based and interpolation-based DA with a computationally inexpensive and straightforward method, SegMix opens up several interesting directions for further exploration.


\section*{Ethics Statement}
Our research does not present any new datasets but rather presents a new general methods that can be used to improve performance of existing NLP applications, and is intended to be used under data-scarce situation. As a result, we anticipate no direct harm involved with the intended usage. However, we realize that it depends on the kind of NLP models/applications the users to apply to. 

Our research does not involve attributing any forms of characteristics to any individual. As a matter of fact, we strive to boost performance for NLP applications on low-resource languages. Our proposed method is easy to implement and adds negligible overhead to computation time compared to similar methods. Due to the fact that we conducted experiments over extensive hyperparameter and data settings, we used around 5000 GPU/hours on Tesla T4 GPUs.

\bibliography{main}
\bibstyle{acl_natbib}

\appendix

\section{Appendix}
\label{sec:appendix}



\subsection{Additional results} \label{sec:add_res}
We conduct experiments on GermEval datasets. The results are included in Table.~\ref{tab:germeval}.
We report the results of the experiment on the varying augmentation rate in MMix, SMix, and TMix in Table~\ref{table:augrate_mmix}. 

\begin{table}[h]
    \centering
    \begin{tabular}{l ccc}\toprule
    & \multicolumn{3}{c}{GermEval} \\ \cmidrule(lr){2-4}
    & $5\%$ & $10\%$ & $30\%$ \\
BERT & $70.28$ & $75.64$ & $79.63$ \\ \midrule
BERT + MR & $74.51$ & $75.98$ & $80.83$ \\
BERT + SR & $73.77$ & $73.26$ & $75.52$ \\
BERT + LR & $73.26$ & $79.49$ & $79.20$ \\ \midrule
BERT + MMix \textdagger & $76.06$ & $80.32$ & $83.48$ \\ 
BERT + SMix \textdagger & $75.07$ & $78.64$ & $80.89$ \\ 
BERT + TMix \textdagger & $74.48$ & $77.07$ & $80.99$ \\ \bottomrule
    \end{tabular}
    \caption{F1 scores on down-sampled GermEval compared with replacement-based augmentation methods. \textdagger denotes our methods.}
    \label{tab:germeval}
\end{table}

\begin{figure}
    \centering
    \includegraphics[width=8cm]{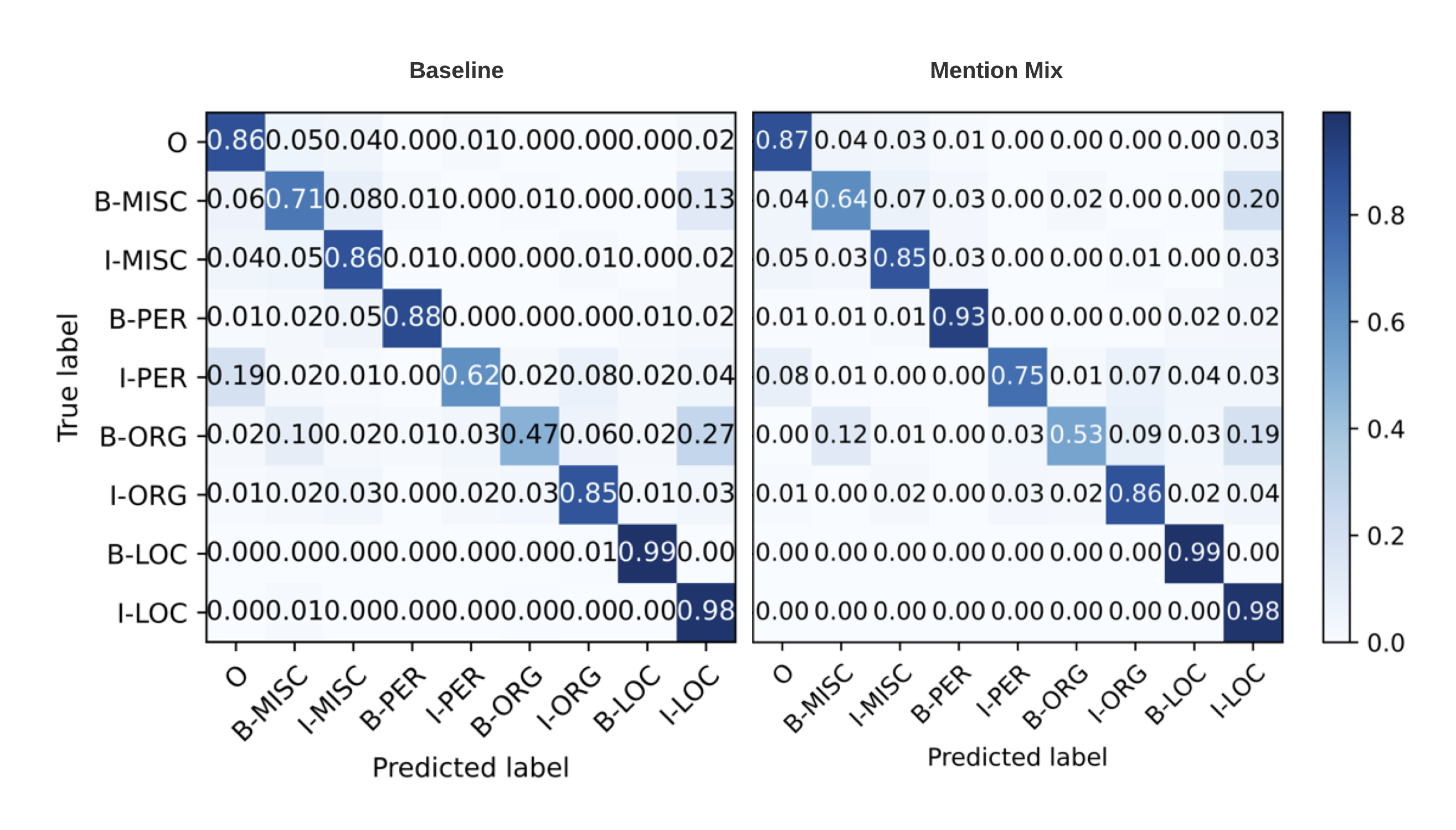}
    \caption{Confusion Matrix on CoNLL-03 with and without SegMix with $200$ training data.}
    \label{fig:confusion_mat}
\end{figure}

To better understand the improvement made by SegMix, we compare the confusion matrix of the baseline model and MMix for each class for $5\%$ of CoNLL-03 data in Fig. \ref{fig:confusion_mat}.

\begin{table}[h]
\centering
\begin{tabular}{l cc}\toprule
Language & Model Link & Reference \\ \midrule
English & \href{https://github.com/huggingface/transformers}{BERT} & \citealt{BERT}\\
English & \href{https://huggingface.co/roberta-base}{RoBERTa} & \citealt{https://doi.org/10.48550/arxiv.1907.11692}\\
Kinyarwanda & \href{https://huggingface.co/Davlan/bert-base-multilingual-cased-finetuned-kinyarwanda}{Kin}&\citealt{10.1162/tacl_a_00416} \\
Sinhala & \href{https://cogcomp.seas.upenn.edu/models/bert_pretrained_models_lorelei/EMBERT_sin_step_500k.tgz}{Sin}& \citealt{wang-etal-2020-extending}\\
\bottomrule
\end{tabular}
\caption{Pre-trained Models}
\label{table:baseline_models}
\end{table}

\begin{table*}[]
\centering
\begin{tabular}{lc|cccccc}\toprule
& \textbf{Aug Rate} & 200 & 400 & 800  & \textbf{Average} \\\midrule
\textbf{Baseline} & 0 & $76.02 \pm 0.56$ & $81.20 \pm 0.29$ & $84.34 \pm 0.33$ & -\\ \midrule
\multirow{6}{*}{\textbf{MMix}}
& 0.1 & $\mathbf{78.76} \pm 0.49$ & $\mathbf{82.28} \pm 0.31$ & $\mathbf{85.51} \pm 0.21$ & $+(1.66 \pm 0.55$)\\
& 0.2 & $77.71 \pm 0.29$ & $82.10 \pm 0.09$ & $84.77 \pm 0.23$ & $+(1.01 \pm 0.47)$\\
& 0.3 & $77.88 \pm 0.20$ & $82.10 \pm 0.19$ & $84.72 \pm 0.28$ &  $+(1.05 \pm 0.47)$ \\
& 0.4 & $77.13 \pm 0.23$ & $81.89 \pm 0.13$ & $84.59 \pm 0.24$ &  $+(0.68 \pm 0.46)$ \\
& 0.5 & $77.38 \pm 0.32$ & $81.32 \pm 0.07$ & $84.66 \pm 0.07$ & $+(0.60\pm 0.47)$ \\ 
& \textbf{Average} & $78.16 \pm 0.44$ & $82.32 \pm 0.26$ & $85.12 \pm 0.17$ & $+(1.00\pm 0.48)$  \\ \midrule
\multirow{6}{*}{\textbf{TMix}} 
& 0.1 & $\mathbf{78.70} \pm 0.47$ & $\mathbf{82.98} \pm 0.27$ & $\mathbf{85.37} \pm 0.26$ & $+(1.83 \pm 0.54$)\\
& 0.2 & $78.51 \pm 0.34$ & $82.35 \pm 0.12$ & $85.26 \pm 0.23$ & $+(1.52 \pm 0.48)$\\
& 0.3 & $78.24 \pm 0.39$ & $82.21 \pm 0.15$ & $85.07 \pm 0.12$ &  $+(1.32 \pm 0.48)$ \\
& 0.4 & $77.56 \pm 0.49$ & $82.11 \pm 0.33$ & $85.22 \pm 0.06$ &  $+(1.11 \pm 0.54)$ \\
& 0.5 & $77.78 \pm 0.60$ & $81.97 \pm 0.17$ & $84.68 \pm 0.25$ & $+(0.96\pm 0.57)$ \\ 
& \textbf{Average} & $78.16 \pm 0.44$ & $82.32 \pm 0.26$ & $85.12 \pm 0.17$ & $+(1.35\pm 0.51)$  \\\midrule
\multirow{6}{*}{\textbf{SMix}} 
& 0.1 & $\mathbf{77.95} \pm 0.39$ & $\mathbf{82.52} \pm 0.36$ & $\mathbf{85.33} \pm 0.19 $ & $+(1.4 \pm 0.52$)\\
& 0.2 & $77.75 \pm 0.46$ & $82.42 \pm 0.35$ & $85.05 \pm 0.18 $ & $+(1.22 \pm 0.54)$\\
& 0.3 & $77.24 \pm 0.44$ & $82.11 \pm 0.07$ & $84.90 \pm 0.16 $ &  $+(0.89 \pm 0.49)$ \\
& 0.4 & $77.23 \pm 0.59$ & $81.75 \pm 0.29$ & $84.76 \pm 0.15 $ &  $+(0.73 \pm 0.57)$ \\
& 0.5 & $77.78 \pm 0.49$ & $81.42 \pm 0.35$ & $84.98 \pm 0.21 $ & $+(0.54\pm 0.55)$ \\ 
& \textbf{Average} & $77.39 \pm 0.50$ & $82.04 \pm 0.29$ & $85.01 \pm 0.17$ & $+(0.96\pm 0.54)$  \\
\bottomrule
\end{tabular}
\caption{f1 scores of MMix, TMix, SMix on CoNLL-03 with variant augmentation rates ($\frac{\# \text{of augmented data}}{\# \text{of training data}}$) under different initial data sizes. SegMix consistently improves over the baseline, demonstrating its stability and robustness over varying augmentation rates. The last row is the averaged improvement score for each augmentation rate over different initial data sizes. The last column is the average score for each initial data size over different augmentation rates.}
\label{table:augrate_mmix}
\end{table*}

\subsection{Variants of BERT Models} \label{subsec:baseline}

As mentioned in Sec.~\ref{subsec:impledetails}, we adopted language-specific BERT models as the pre-trained models for all tasks. There are 12 layers (transformer blocks), 12 attention heads, and 110 million parameters \citep{BERT}. The model links are included in Table.~\ref{table:baseline_models}.
For Kinyarwanda, \textit{bert-base-multilingual-cased-finetuned-kinyarwanda} is obtained by fine-tuning Multilingual BERT (MBERT) on the Kinyarwanda dataset JW300, KIRNEWS, and BBC Gahuza ~\citep{10.1162/tacl_a_00416}. \textit{EMBERT-Sin} is obtained by EXTEND~\citep{wang-etal-2020-extending} MBERT in Sinhala. Specifically, \textit{EMBERT-Sin} first incorporates the target language Sinhala by expanding the vocabulary, and then continues pre-training on LORELEI using a batch size of $32$, a learning rate of $2e-5$, and trained for $500K$ iterations.

\begin{table*}[]
    \centering
    \begin{tabular}{l | l | p{11.5 cm}}\toprule
        \multirow{2}{*}{Pred. 1} & Baseline  &\textcolor{blue}{\textbf{English}} \scriptsize\textcolor{blue}{[MISC]} \normalsize county sides and another against \textcolor{red}{\textbf{British Universities}} \scriptsize\textcolor{red}{[MISC]}
        \normalsize \\
        & MMix  & \textcolor{blue}{\textbf{English}} \scriptsize\textcolor{blue}{[MISC]} \normalsize county sides and another against \textcolor{blue}{\textbf{British Universities}} \scriptsize\textcolor{blue}{[ORG]}
        \normalsize \\\midrule
        \multirow{2}{*}{Pred. 2} & Baseline  & May 22 First one-day international at \textcolor{red}{\textbf{Headingley}} \scriptsize\textcolor{red}{[ORG]} \normalsize \\
        & MMix  & May 22 First one-day international at \textcolor{blue}{\textbf{Headingley}} \scriptsize\textcolor{blue}{[LOC]} \normalsize \\\midrule
        \multirow{2}{*}{Pred. 3} &Baseline  & July 9 v \textcolor{red}{\textbf{Minor Counties}} \scriptsize\textcolor{red}{[MISC]} \normalsize XI\\
        & MMix  & July 9 v \textcolor{red}{\textbf{Minor Counties}} \scriptsize\textcolor{red}{[ORG]} \normalsize XI\\
        \bottomrule
    \end{tabular}
    \caption{Examples of cases predicted by the baseline model and MMix from validation dataset. The colored segments represent an entity mention, the blue segment represents a correctly classified mention, and the red represents a misclassified mention.}
    \label{tab:error_cases}
\end{table*}

\begin{table*}[]
    \centering
    \begin{tabular}{l|l|l}\toprule
        Ex. 4 & \multicolumn{2}{p{12cm}}{the complete [\textbf{statue}]\small\textcolor{red}{$e_1$}\normalsize  ~topped by an imposing [\textbf{head}]\small\textcolor{red}{$e_2$}\normalsize was originally nearly five metres high} \\
        \textbf{Other}  &
        Baseline:  \textcolor{blue}{Component-Whole(e2,e1)} &
        RMix : Other\\\midrule
        Ex. 5 & \multicolumn{2}{p{12cm}}{the [\textbf{slide}]\small\textcolor{red}{$e_1$}\normalsize which was triggered by an avalanche - control [\textbf{crew}] \small\textcolor{red}{$e_2$}\normalsize ~damaged one home and blocked the road for most of the day } \\
        \textbf{Cause-Effect(e2,e1)} &
        Baseline:  \textcolor{blue}{Product-Producer(e1,e2)} &
        RMix : \textcolor{blue}{Cause-Effect(e1,e2)}\\
        \bottomrule
    \end{tabular}
    \caption{Examples of correctly classified cases after RMix. The bold segment tuple represents a nominal pair, and the blue label represents a misclassified relation. The true label is presented in the first column.}
    \label{tab:error_cases_rela}
\end{table*}

\subsection{Case Analysis} \label{subsec:case_analysis}
We list some improved cases in Table.~\ref{tab:error_cases}, Ex. 1 and 2 are cases of correction between for ORG, while Ex. 3 is a case where the entity label is correct, but the mention range remains incomplete~(both predicts \textit{Minor Counties} as a mention instead of \textit{Minor Counties XI}).
In Table.~\ref{tab:error_cases_rela}, we list some improved cases for RMix on RE. Both Ex.4 and 5 are cases of correction for relation type. In Ex.5, RMix helps the model classify the correct relation but not in the correct order.

\end{document}